\documentclass[lettersize,journal]{IEEEtran}

\usepackage{amsmath,amssymb}
\usepackage{graphicx}
\usepackage{algorithm}
\usepackage{algpseudocode}
\usepackage{tabularx}
\usepackage{multirow}

\usepackage{tikz}
\usetikzlibrary{matrix, arrows.meta, positioning}

\begin{document}

\title{SPIN: Decentralized Swarm Organization via Tensorized Policy Coordination}

\author{Zhaowen~Fan, 
        and~Yunxiang~Han%
\thanks{Zhaowen Fan and Yunxiang Han are with the School of
Computer Science, Sichuan University, Chengdu 610065, China (e-mail:
iamsuperfan@stu.scu.edu.cn; hanscu@scu.edu.cn).}%
\thanks{Corresponding author: Yunxiang Han.}}

\maketitle

\begin{abstract}
Decentralized multi-agent swarm coordination remains fundamentally challenged by the
combinatorial scaling of joint action spaces and high-overhead training or
optimization loops when managing localized group behaviors. To address this problem
from a different perspective, this paper introduces the Swarm Policy Interference
Network (SPIN) framework, which models multi-agent communication topologies as
compressed tensor networks. By factorizing the joint policy tensors of local
multi-agent cliques into Open Boundary Condition Matrix Product State (MPS) chains,
SPIN replaces explicit exponential joint-action enumeration with clique-level
contractions that scale linearly in clique length for fixed local behavior and
bond dimensions. To connect raw spatial geometry with this discrete algebraic backend
without relying on online training loops, we introduce a decoupled framework combining
a lightweight frozen neural mapper evaluated offline with a deterministic zero-shot
importance-reweighting filter based on the Radon-Nikodým derivative. We evaluate an
executable prototype of this framework within a simulation experiment across distinct
task regimes: single-target tracking, decentralized area coverage, and structured
multi-goal coordination. The results demonstrate that SPIN functions as a reusable
decentralized coordination layer across tracking, dispersion / area coverage, and
structured multi-goal coordination, with its strongest gains appearing in multi-goal
coordination and dense local interaction regimes, without requiring scenario-specific
online optimization or retraining.
\end{abstract}

\section{Introduction}

Autonomous robotic swarms, including micro-Unmanned Aerial Vehicles (UAVs), present
a transformative opportunity for distributed target tracking, environmental
monitoring, and decentralized search-and-rescue operations \cite{sheng2006distributed,
chung2006decentralized}. More broadly, swarm robotics has been framed as a problem of
swarm engineering, that is, the systematic design, analysis, validation, and operation
of decentralized multi-robot systems whose global behavior emerges from localized
interactions \cite{brambilla2013swarm}. However, to the best of our knowledge, a
central unresolved difficulty remains: there is still no universal and systematic
method for characterizing global collective behavior directly from simple local agent
rules, especially as swarm size and interaction complexity increase
\cite{brambilla2013swarm}. This challenge becomes particularly acute in UAV settings,
where coordination must remain decentralized under intermittent ad hoc connectivity
and limited onboard resources \cite{bekmezci2013flying, hayat2016survey}.

Two core structural bottlenecks continue to limit the design of expressive
multi-agent coordination architectures. First of all, there is the so-called
dimensionality problem inherent to joint action spaces. In a system where $m$
neighboring agents must coordinate across $n$ discrete macro-behaviors, the
corresponding joint probability tensor scales exponentially as $O(n^m)$
\cite{rashid2020monotonic}. Even when such structures are represented only locally,
explicit evaluation or update of coupled multi-agent state tables can rapidly become
unwieldy as local interaction density grows. To mitigate this, many structured
probabilistic approaches rely on iterative consensus, message-passing, or
belief-propagation-style coordination \cite{pearl2014probabilistic,
ihler2004nonparametric}. While such mechanisms can preserve distributed structure,
they also introduce additional coordination overhead and may become fragile under
noise, packet loss, or partial observability.

Secondly, the runtime adaptation of policies from continuous sensorimotor inputs is
also a concern. Real-world swarms must convert local geometric perception streams,
such as relative position estimates or richer sensing modalities including
LiDAR-derived observations \cite{li2020deep}, into stable collective control
behavior. Large-scale multi-agent reinforcement learning has achieved impressive
results in complex simulated coordination domains \cite{vinyals2019grandmaster},
transferring such high-capacity learned policies directly to decentralized swarm
settings remains difficult. Deep neural networks are powerful function approximators,
yet architectures that depend on repeated online backpropagation or heavy stochastic
optimization during execution are difficult to reconcile with decentralized swarm
settings \cite{mozaffari2019tutorial, svoboda2022deep}. These limitations
motivate coordination frameworks that avoid both explicit combinatorial joint-policy
representations and costly online learning loops, while still retaining sufficient
structure to model localized multi-agent dependence.

Modern swarm-intelligence literature has increasingly explored hybrid decentralized
architectures that combine learned local policies with structured coordination
priors \cite{li2017learning, battaglia2018relational}, this trend aligns with a
broader recent interest in swarm-intelligence methods that emphasize structured local
interaction, robustness under communication constraints, and reduced reliance on
centralized optimization \cite{brambilla2013swarm}. Rather than drifting with the
tide, we propose the Swarm Policy Interference Network (SPIN) framework, a
decentralized coordination framework that combines clique-aware tensor compression
with zero-shot policy reweighting. The representational design of SPIN draws on
quantum-formalism-inspired notation only in a limited algorithmic sense, and is not
intended as a physical quantum model. Related quantum-inspired optimization paradigms
\cite{busemeyer2012quantum} are often formulated for offline metaheuristic search or
abstract static decision settings \cite{sun2004particle}. In SPIN, this influence is
limited to the representational level, namely the use of complex-compatible local
state vectors, clique-level tensor factorizations, and marginal-recovery operators for
decentralized coordination. Additionally, its behavior is driven by bounded
likelihood-ratio reweighting, clique-consistent tensorized coordination, and a
mapping to classical motion commands, yielding a structured alternative to purely
end-to-end learned multi-agent coordination pipelines \cite{verbraeken2020survey}.

Unlike learned communication architectures in graph-neural-network-based MARL,
where coordination emerges through parameterized message passing
\cite{foerster2016learning, das2019tarmac}, SPIN treats coordination as an explicit
executable mechanism operating over structured local interactions. Furthermore,
SPIN does not aim to compress a learned joint policy or value function
\cite{yu2022surprising, rashid2020monotonic}; instead, tensorized representations are
used to organize localized clique-level coordination states, while runtime adaptation
is achieved through bounded likelihood-ratio reweighting and marginal reconciliation
without online policy optimization \cite{schulman2015trust, schulman2017proximal}.
Therefore, the tensor-network component should be viewed as a structured
representation substrate rather than the primary contribution itself. The central
contribution of SPIN lies in enabling structured and computationally bounded
decentralized coordination under localized interaction complexity.

Instead of modeling decentralized coordination as a sequence of repeatedly optimized
local policies, we formulate swarm coordination as a structured probabilistic
inference problem over a time-varying communication graph
\cite{koller2009probabilistic, wainwright2008graphical}. The communication topology
is represented as a Markov Random Field whose localized clique interactions are
compressed through tensor-network factorizations \cite{orus2014practical}, while
geometric perception and behavioral adaptation are explicitly separated into
independent computational stages. This formulation enables lightweight perception,
structured coordination, and deterministic runtime adaptation to operate as
complementary components within a unified decentralized coordination framework. The
proposed framework integrates three key conceptual ideas: (i) tensor-network policy
compression for clique-level coordination under growing local interaction complexity,
(ii) neuro-symbolic perception that separates offline geometric representation
learning from runtime behavioral modulation, and (iii) clique-aware tensorized policy
inference through bounded likelihood-ratio reweighting and marginal projection.
Together, these components provide a unified perspective on decentralized perception
and coordination for swarm behavior.

To the best of our knowledge, SPIN is among the first decentralized swarm coordination
frameworks to jointly integrate tensor-network policy compression, offline
neuro-symbolic perception, and clique-aware probabilistic coordination within a
unified decentralized architecture. Experimental results demonstrate that the proposed
framework admits an executable structured coordination mechanism that mitigates
explicit local joint-policy explosion at the representation level, supports multiple
collective task regimes within a shared pipeline, and exhibits task-dependent
strengths, particularly in multi-goal coordination, together with measured robustness
under controlled noise and information-dropout stress tests. The paper further
includes a focused component-level validation of the proposed tensor-compression
mechanism through direct comparison with explicit joint-state enumeration and scaling
measurements under increasing clique size. The remainder of this paper is organized
as follows. Section~II presents the methodology, including the localized state
representation, tensor-compressed clique factorization, overlap-consistency
constraints, bounded likelihood-ratio reweighting, and the momentum-attenuation
discussion. Section~III describes the algorithmic implementation and simulation
pipeline, including the synchronous PettingZoo-compatible control loop, perceptual
pre-training, topology handling, and baseline setup. Section~IV defines the system
evaluation setting and protocol. Section~V reports the experimental results,
including observed behavioral regimes, baseline comparisons, repeated-trial
summaries, tensor-compression validation and scalability, and perception robustness.
Section~VI discusses the results and current evaluation limitations. Section~VII
concludes with final remarks and future directions.

\section{Methodology}

\subsection{The Local State Space and Tensorized Coordination Mapping}

To maintain a low computational complexity per agent, we define a Localized 
State Subspace as the minimum partition of the global swarm state assigned 
to individual drones and compressed via tensor networks. Each drone $i$ 
maintains a local latent coordination state vector $|\psi_i\rangle$ as a
collection of latent behavioral feature bases $\{|s_1\rangle, |s_2\rangle, 
\dots, |s_n\rangle\}$ defined by the target mission's event abstract:

\begin{equation}
    |\psi_i\rangle = \sum_{k=1}^{n} \alpha_{i,k} |s_k\rangle,
    \quad \alpha_{i,k} \in \mathbb{C}
\end{equation}

where the coefficient $\alpha_{i,k}$ is carried in a complex-compatible
representation inside the tensorized coordination layer, while the executable
agent policy depends on normalized magnitudes and marginalized clique summaries
rather than on persistent agent-level complex-state trajectories. Target
activation probabilities are obtained by squared-magnitude normalization:

\begin{equation}
    P_i(s_k) = |\alpha_{i,k}|^2, \quad \text{subject to} \quad \sum_{k=1}^{n}
    |\alpha_{i,k}|^2 = 1
\end{equation}

Rather than directly determining motion on their own, these derived internal
policy weights ($P_i(s_k)$) function as high-level symbolic modulation coefficients.
At the execution layer, the agent's executable continuous action is computed by
mapping the local behavior probabilities together with the continuous policy signal
into the simulator's five-dimensional action interface. This structural
decoupling ensures that swarm-level coordination is driven by compressed algebraic
tensor interactions, while motion execution remains bounded through simple geometric
update rules and simulator-side safeguards.

SPIN should therefore be interpreted as a structured coordination layer rather
than as a complete low-level controller. The internal complex-valued tensorized state
is not itself the actuator, instead, it serves as a compact representation of
clique-level coordination structure. The executed motion remains real-valued and
bounded, but the weights driving that motion are first shaped by a localized
multi-agent algebraic coordination stage. In this sense, the final motion map does
not discard the internal tensor computation, but rather operationalizes it through
a bounded continuous control interface.

To evaluate behavioral adaptation dynamically without introducing geometric
initialization priors or pre-exposure advantages, the local coordination weights
are initialized to a uniform maximum-entropy prior. This establishes a strict
baseline distribution across all behavioral elements prior to any sensor-driven
Radon-Nikodým filtering. This uniform initialization ensures that subsequent
swarm coordination patterns emerge entirely from runtime environmental
interactions rather than from initialization artifacts.

\subsection{Tensor-Compressed Markov Random Field Cliques}

The communication topology of the swarm is modeled as an undirected, time-varying
Markov Random Field (MRF) $\mathcal{G} = (\mathcal{V}, \mathcal{E})$ over a
decentralized ad hoc network topology. Rather than restricting agents to
localized line-of-sight visual perception, the current simulation prototype
assumes that each node $\mathcal{V}$ ingests shared relative-coordinate state
arrays representing the relative positions of peer drones and environmental
landmarks. These arrays are aggregated onboard into local network coordinate
matrices. To preserve bounded local computation under crowded deployments, each
agent retains at most $k_{\max}=8$ nearest neighbors within the sensing radius
$R_{\text{sense}}$ before clique extraction. This converts the raw radius graph
into a capped local interaction graph, allowing the swarm to dynamically form
localized, maximal communicative cliques $\{C_A, C_B, \dots\} \subset
\mathcal{G}$ without requiring a centralized base station or global optimization
server.

To bypass the curse of dimensionality inherent to joint multi-agent action
spaces, where tracking a standard joint probability tensor scales exponentially
as $O(n^m)$ for a clique of size $m$, we factorize the joint clique policy tensor
$|\Psi_{C_A}\rangle$ into a localized Matrix Product State (MPS) chain
(tensor-train factorization):

\begingroup
\small
\begin{equation}
    |\Psi_{C_A}\rangle = \sum_{s_1, \dots, s_m} A_1[s_1] A_2[s_2] \dots A_m[s_m]
    |s_1, s_2, \dots, s_m\rangle
\end{equation}
\endgroup

where each $A_i[s_i]$ is a local tensor core. To preserve efficient contraction
during execution, we enforce open boundary conditions (OBC) rather than periodic
loop matrices: the boundary cores $A_1[s_1]$ and $A_m[s_m]$ are explicitly
constrained to row and column vectors of dimension $1 \times \chi$ and
$\chi \times 1$ respectively, while internal cores maintain a matrix dimension of
$\chi \times \chi$. This open boundary chain eliminates periodic closure and yields
contraction cost that is linear in clique length when the local behavior dimension
and bond dimension are treated as fixed. The neighbor cap does not change the
tensor-network contraction rule inside a given clique. Instead, it bounds the
practical size of the local interaction graph from which cliques are extracted.
Thus, the MPS reduction still applies to each retained clique, while $k_{\max}$
prevents dense 2D deployments from creating large radius-induced cliques that
dominate runtime.

In the current implementation, spatial dispersion and anti-collapse bias enter
through the continuous policy signal, where task-directed geometric drive is combined
with a local repulsion vector before the frozen neural mapping stage. Accordingly,
geometric repulsion influences the downstream target measure, local state update,
and final bounded action map before clique-wise marginalization, without requiring
explicit online state negotiation.

\subsection{Partial Trace Consistency Constraints}

In the absence of a centralized coordinator, cross-clique coordination coherence is
maintained by leveraging overlapping MRF cliques as localized synchronization bridges.
If an individual agent $i$ simultaneously occupies overlapping cliques $C_A$ and
$C_B$, it serves as a structural link for state propagation. To prevent systemic
divergence, the framework enforces an algebraic consistency constraint requiring
the agent-local marginal operator $\rho_i$ of the agent to be invariant
regardless of its parent clique context:

\begin{equation}
    \rho_i = \text{Tr}_{\backslash i} (|\Psi_{C_A}\rangle\langle\Psi_{C_A}|) =
    \text{Tr}_{\backslash i} (|\Psi_{C_B}\rangle\langle\Psi_{C_B}|)
\end{equation}

where $|\Psi_{C_A}\rangle$ and $|\Psi_{C_B}\rangle$ denote the joint clique
states of cliques $C_A$ and $C_B$, and where $\text{Tr}_{\backslash i}$
denotes exact tensor marginalization (partial trace) over all clique variables
excluding agent $i$. While classical multi-agent graphical models typically
enforce sub-graph consistency through high-latency iterative belief propagation
or consensus message-passing, this formulation computes localized summaries
algebraically via direct tensor-train contractions.

To account for possible asynchronous or incomplete local coordination contexts, this
strict consistency condition is relaxed online into an iterative tracking objective.
Agent $i$ monitors its structural divergence across overlapping contexts by
evaluating the trace distance discrepancy:

\begin{equation}
    \mathcal{L}_{\text{sync}}(\rho_i) = \frac{1}{2} \left\|
    \text{Tr}_{\backslash i} (|\Psi_{C_A}\rangle\langle\Psi_{C_A}|) -
    \text{Tr}_{\backslash i} (|\Psi_{C_B}\rangle\langle\Psi_{C_B}|) \right\|_1
\end{equation}

Executing continuous semidefinite programming (SDP) or centralized non-linear
optimization to minimize $\mathcal{L}_{\text{sync}}$ at every timestep is
computationally expensive for the intended decentralized execution setting. To
satisfy real-time, high-frequency execution deadlines, this continuous minimization
is approximated via an iterative trace-distance consensus update over overlapping
clique contributions.

By executing left-right environment contractions directly within the open-boundary
Tensor-Train cores, the implementation computes clique-local reduced densities and
then reconciles overlapping clique contributions at the agent level. This structural
relaxation performs an approximate trace-distance consensus update across overlapping
clique contributions while avoiding the cost of active online matrix optimization
loops. The tensor-network layer thus acts not to replace classical motion primitives,
but to reshape them through a compact local coordination envelope, ensuring
overlapping agents are influenced by shared clique context before any physical motion
command is issued.

\subsection{Bounded Likelihood-Ratio Reweighting (Radon-Nikodým Form)}

To execute instantaneous behavioral adaptations without relying on power-intensive
online training loops, SPIN decouples raw spatial signal processing from
downstream algebraic tensor coordination. Rather than maintaining independent
neural weight matrices per agent, the swarm utilizes a single, parameter-reused
coordination mapping network $\phi_\omega$ instantiated globally and evaluated
locally on agent-centric features. For an individual agent $i$ at time step $t$,
handcrafted geometric steering laws first aggregate local relative target
coordinates and proximity-based repulsion vectors into a continuous 2D spatial
input descriptor, $\text{signal}_t^i$. The parameter-frozen network processes this
composite spatial descriptor in a single zero-shot forward pass to generate a
target behavioral distribution:

\begin{equation}
    \nu_i = \phi_\omega(\text{signal}_t^i)
\end{equation}

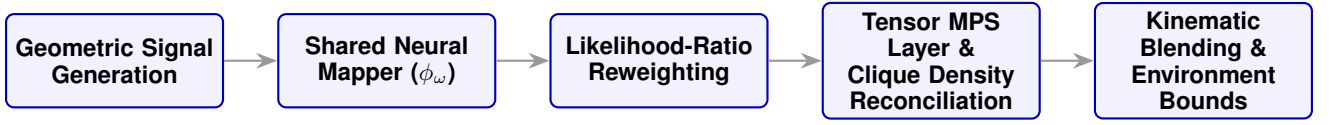
\begin{figure*}[ht]
\centering
\begin{tikzpicture}[
    block/.style={
        draw, 
        rectangle, 
        rounded corners=3pt,
        thick,
        fill=blue!5, 
        draw=blue!70!black,
        text centered, 
        minimum height=3.5em,
        text width=7.5em, 
        font=\small\sffamily\bfseries
    },
    arrow/.style={
        ->, 
        >={Stealth[scale=1.2]}, 
        thick, 
        draw=gray!80
    }
]

\matrix (m) [matrix of nodes, column sep=2.0em, row sep=1em, nodes={anchor=center}] {
    |[block]| {Geometric Signal\\Generation} &
    |[block]| {Shared Neural\\Mapper ($\phi_\omega$)} &
    |[block]| {Likelihood-Ratio\\Reweighting} &
    |[block]| {Tensor MPS Layer \&\\Clique Density\\Reconciliation} &
    |[block]| {Kinematic Blending \&\\Environment Bounds} \\
};

\draw [arrow] (m-1-1) -- (m-1-2);
\draw [arrow] (m-1-2) -- (m-1-3);
\draw [arrow] (m-1-3) -- (m-1-4);
\draw [arrow] (m-1-4) -- (m-1-5);

\end{tikzpicture}
\caption{SPIN Operational Control Pipeline Architecture.}
\label{fig:SPIN_pipeline}
\end{figure*}

This explicit pipeline architecture maps continuous local geometric inputs directly
into the discrete algebraic coordination layer, establishing the complete execution
topology detailed in Figure~\ref{fig:SPIN_pipeline}.

To formalize the underlying multi-agent tensor structure, each agent's localized
behavioral profile is tracked via a normalized state ket $|\psi_i\rangle =
\sum_{k=1}^n \alpha_{i,k} |s_k\rangle$ defined over a finite, orthogonal behavioral
action basis $\{|s_k\rangle\}_{k=1}^n$. Crucially, the coefficients $\alpha_{i,k} \in
\mathbb{C}$ are complex-valued, allowing the downstream tensor tracking layers
to execute phase modulations and multi-agent state reductions. Let $\mathbf{\mu}_i
\in \mathbb{R}^n$ represent the baseline behavioral probability vector recovered
from the prior state magnitudes via a Born-rule projection ($P_i(s_k) =
|\alpha_{i,k}|^2$), and let $\mathbf{\nu}_i \in \mathbb{R}^n$ define the target
distribution generated by $\phi_\omega$. In this representation, squared magnitudes
determine executable behavioral probabilities, while phase is used only as an
internal structural degree of freedom for tensorized coordination updates.

Over this discrete action basis, the implemented likelihood ratio is evaluated
element-wise using a small probability floor, namely $\left[\frac{d\nu_i}{d\mu_i}
\right]_k \approx \frac{\nu_i(s_k)}{\max(\mu_i(s_k), 10^{-8})}$, before square-root
scaling and gain clipping. At runtime, SPIN leverages this density shift to construct
a continuous, non-unitary policy driving operator $\hat{M}(\Delta t)$ that modulates
the magnitudes of the complex state components:

\begin{equation}
    \hat{M}(\Delta t) = \exp\left( \left[ \text{diag}\left(\min\left(
    \sqrt{\frac{d\nu}{d\mu}}, \gamma\right)\right) - \mathbf{I} \right] \Delta
    t \right)
    \label{eq:M_definition}
\end{equation}

where $\gamma$ clips large element-wise gains after the probability floor has been
applied, and $\mathbf{I}$ establishes an equilibrium tracking baseline. The
square-root operator ensures algebraic consistency with the downstream quadratic Born
mapping. Because $\hat{M}(\Delta t)$ acts as an exponential density shift scaled by
the chronological interval $\Delta t$, state components that align with
environmental incentives are amplified smoothly, while unselected pathways
decay continuously. This zero-shot algebraic calculation is evaluated at runtime
without any online backpropagation.

Following this localized reweighting step, the updated state vectors are
passed directly to the tensor network layer. Here, the joint policy states of
localized interaction cliques are factorized into Matrix Product State (MPS)
chains. Because agents can simultaneously participate in intersecting sub-graphs,
the tensor layer executes an overlapping-clique density reconciliation routine.
This step enforces approximate algebraic consistency across shared variables through
an iterative trace-distance consensus update before projection back to the local
amplitude state, without requiring global communication overhead.

To preserve bounded execution and numerical stability under discrete-time
simulation loops, the system bypasses idealized continuous motion equations in
favor of a multi-stage physical execution pipeline:

\begin{enumerate}
    \item \textbf{Nonlinear State Renormalization:} Because the driving operator
    $\hat{M}(\Delta t)$ is non-unitary, it does not intrinsically preserve the
    $L_2$ norm of the state vector. The driven state ket must pass through an
    immediate normalization step to re-establish a valid verification mapping:

    \begin{equation}
        |\psi_{\text{new}}\rangle = \frac{\hat{M}(\Delta t)|\psi_{\text{old}}
        \rangle}{\| \hat{M}(\Delta t)|\psi_{\text{old}}\rangle \|_2}
    \end{equation}

    The realizable action propensities are subsequently recovered from the
    renormalized local state via $P_i(s_k) = |\alpha_{i,k}|^2$.
    
    \item \textbf{Kinematic Blending and Environment-Side Boundaries:} The final
    execution commands are produced by mapping the Born probabilities together with
    the continuous policy signal onto the simulator's five-dimensional continuous
    action convention. Rather than evaluating a single analytical continuous force
    equation, collision avoidance and workspace constraints are distributed across
    decoupled structural layers. Local inter-agent repulsion is embedded natively
    within the initial spatial features feeding the network, while physical arena
    boundaries ($\Omega$) are monitored directly by the environment wrapper. When an
    agent encounters a workspace perimeter, the environment executes a deterministic
    velocity clipping and an elastic bounce-damping mechanism. This multi-layered
    separation ensures that abstract high-level selection matches the bounded
    execution constraints imposed by the simulation environment.
\end{enumerate}

\subsection{Asymptotic Convergence and Momentum Attenuation}

While the non-unitary reweighting formulation proposed in
Equation~\ref{eq:M_definition} eliminates the pathological zero-shot degenerate
policy concentration characteristic of discrete updates, it introduces a
continuous behavioral momentum parameterized by $\Delta t$. Empirical simulation
testing reveals that under conditions of extreme initial topological isolation
(clique size $|C_m| = 1$), an agent operating purely on isolated frozen-forward
states can accumulate localized amplitude bias faster than spatial convergence
can dictate. 

To prevent runaway kinetic trajectories in unbound coordinate topologies, an
explicit spatial clamping boundary constraint is enforced during the hybrid
kinematic deployment step, ensuring that positions are strictly projected back
into the operational domain: $\Omega \in [0, 100]^2$. This safeguard prevents
unbounded positional divergence within the finite simulation domain even when an
agent remains temporarily isolated.

\section{Algorithmic Implementation and Simulation}

It is important to argue that the current SPIN framework is necessarily focused
on the coordination layer of the swarm behavior, therefore the principle of the
experiment setting is to isolate the coordination layer from task-specific lower-level
control. This design improves comparability and reproducibility, making future
coupling with other low-level controllers easier. To instantiate and evaluate the SPIN
coordination mechanism, we use a bounded two-dimensional simulation with $N=10$ agents
over $T=120$ synchronized control intervals. Each episode starts from a random
non-overlapping placement of the agents inside the $100 \times 100$ arena. The
environment exposes a PettingZoo-style parallel interface, but it does not use dense
reward shaping: all scalar rewards are fixed at zero. The observed behavior therefore
comes entirely from the zero-shot control pipeline rather than from reward-driven
learning. To preserve bounded and comparable kinematic execution across all
controllers, the simulator also applies a post-step body-exclusion correction that
separates interpenetrating agents and damps their velocities whenever pairwise overlap
is detected.

\subsection{Core Control Loop Execution Pipeline}

At each control interval, the simulator executes four stages: adjacency construction
from pairwise Euclidean distances, extraction of overlapping maximal cliques,
clique-wise state representation inference, and bounded continuous motion execution.
The local behavioral basis contains five macro-actions: \emph{north}, \emph{south},
\emph{east}, \emph{west}, and \emph{pinpoint}. A compact offline-trained MLP maps each
agent's control signal to a normalized target measure, and its parameters remain
frozen during execution. Algorithm~\ref{alg:SPIN_loop} summarizes the full synchronous
control loop.

\begin{algorithm}[t]
\footnotesize
\caption{SPIN Synchronous Edge Control Loop}
\label{alg:SPIN_loop}
\begin{algorithmic}[1]
\State \textbf{Input:} Initial positions $\vec{x}$, sensing radius
$R_{\text{sense}}$, neighbor cap $k_{\max}$, clamp $\gamma$, shared frozen
network $\phi_{\omega}$, horizon $T$, scenario mode.
\State \textbf{Output:} Agent trajectories and metrics.
\State Initialize local weights to uniform prior $|\psi_i(0)\rangle$

\For{$t = 1$ \textbf{to} $T$}
    \State Construct radius graph from pairwise distances using $R_{\text{sense}}$
    \State Prune each node to at most $k_{\max}$ mutual nearest neighbors
    \State Extract maximal cliques $\{C_m\}$ from the capped local interaction graph

    \For{each agent $i$}
        \State Compute agent-centric spatial signal $\text{signal}_t^i$
        \State Evaluate target measure $\nu_i = \phi_\omega(\text{signal}_t^i)$
        \If{scenario is Dispersion / Area Coverage}
            \State Apply coverage-aware modulation to $\nu_i$
        \EndIf
        \State Compute bounded likelihood-ratio gain from $\frac{d\nu_i}{d\mu_i}$
        \State Apply non-unitary reweighting and renormalize $|\psi_i\rangle$
    \EndFor
    
    \For{each clique $C_m$}
        \State Build clique-local MPS with mode-dependent phase structure
        \State Compute one-site reduced marginal densities $\rho_i$
    \EndFor

    \For{each agent $i$ in intersecting cliques}
        \State Reconcile overlapping marginal contributions via iterative
        trace-distance-based density consensus
    \EndFor
    
    \For{each agent $i$}
        \State Compute action probabilities $P_i(s_k) = |\alpha_{i,k}|^2$
        \State Map the Born probabilities together with the continuous policy signal
        into environment action commands
        \State Apply environment-side velocity damping, speed clipping, and
        boundary handling
    \EndFor
    
    \State Step environment and record metrics
\EndFor
\end{algorithmic}
\end{algorithm}

An important implementation detail is that the final action passed to the simulator
is always a bounded real-valued motion command. This projection should not be
interpreted as collapsing the internal coordination representation into a trivial
heuristic. Rather, it serves as the execution interface between the clique-aware
coordination state and bounded motion in the continuous arena. The tensorized internal
layer converts those weights into bounded kinematic commands.

From an implementation perspective, the execution layer of SPIN is governed by a small
set of explicit runtime safeguards rather than by a formal global stability analysis.
In particular, the simulator enforces bounded motion through a fixed action scale,
velocity damping, and a maximum speed cap; agent positions are projected back into the
finite arena after each step; and a post-step overlap-correction rule separates
interpenetrating agents while damping their local velocities. These mechanisms do not
constitute a formal control-barrier or Lyapunov-style stability analysis, but they
ensure that the reported trajectories are generated under bounded, overlap-aware
kinematic execution consistent with the released implementation.

In the Dispersion / Area Coverage regime, coverage-aware modulation denotes the
deterministic adjustment of the target measure toward spreading and local
de-crowding before bounded likelihood-ratio reweighting. In the implementation,
this adjustment modifies the measure passed into the reweighting stage rather
than introducing an additional learned module or a separate task-specific
controller.

\subsection{Perceptual Pre-training and Spatial Biasing}

To eliminate runtime learning overhead, the Multi-Layer Perceptron (MLP)
parameterizing $\phi_\omega$ is trained offline using synthetic data over $E = 5000$
epochs. During pre-training, uniform spatial coordinates are sampled to construct an
observation vector $\vec{o} = \vec{x}_{\text{target}} - \vec{x}_{\text{drone}}$.
To avoid directional cancellation during continuous vector blending, the supervised
ground-truth target measure $\nu_{\text{true}}$ is dynamically biased based on the
dominant quadrant of the tracking vector:

\begin{equation}
    \nu_{\text{true}} = 
    \begin{cases} 
    [0.05, 0.05, 0.05, 0.05, 0.80]^T & \text{if } \|\vec{o}\| < 20.0 \\
    \text{Bias}(\text{arg max}_{d \in \{x,y\}} |o_d|) & \text{if } \|\vec{o}\|
    \ge 20.0
    \end{cases}
\end{equation}

The analytical gradients are propagated via Cross-Entropy loss backpropagation,
freezing the network parameters $\omega$ prior to deployment. At runtime, the
agent performs a single zero-shot forward pass with stable Softmax normalization
to yield the environment demand vector.

\subsection{Vectorized Ad Hoc Topology and Zero-Shot Algebraic Reweight-Filtering}

Inter-agent networking avoids standard graph traversal bottle-necks by maintaining
high-speed broadcasting operations. The localized interaction graph is constructed
through synchronized pairwise distance evaluation:

\begin{equation}
    \mathbf{D}_{i,j} = \|\vec{x}_i - \vec{x}_j\|^2_2
\end{equation}

The continuous distance matrix is first filtered by using the sensing boundary
($R_{\text{sense}} = 15.0$). Under crowded deployments, each node then retains at
most $k_{\max}=8$ mutual nearest neighbors before maximal clique extraction. For each
retained clique, the simulator builds a compact MPS whose local cores encode
mode-dependent state coupling correlations, and computes exact one-site reduced
density matrices using left-right environment contractions rather than explicit
enumeration of the full joint tensor.

The zero-shot Radon-Nikodým transformation then reweights local action amplitudes
while preserving the local matrix-form state representations. The operational
likelihood gain incorporates a clipping safeguard $\gamma = 5.0$ after applying a
small probability floor when the prior measure approaches zero ($\mu \to 0$).
Continuous updates are bounded through final nonlinear state renormalization, with
fallback to a uniform maximum-entropy prior if local state configurations approach
numerical instability. After clique-wise reduced density matrices are obtained, agents
that belong to multiple cliques reconcile their shared local marginal contributions
before projection back to executable local state vectors for continuous motion
execution.

\subsection{Baseline Implementation}

To contextualize SPIN against both classical and learned alternatives, we implement
three matched baselines in the same setting. The APF-Velocity baseline is a
deterministic artificial-potential-field controller combining linear target
attraction, inverse-cube inter-agent repulsion, and a soft inward boundary field. The
Distributed Auction-CBBA baseline constructs ring slots around active targets or
coverage anchors and uses a lightweight auction-style assignment procedure to allocate
agents to those slots before applying local repulsion and boundary regularization. The
MAPPO baseline is instantiated from the PPO-based cooperative multi-agent
reference implementation of Yu et al.~\cite{yu2022the}. Unlike SPIN, it is
trained separately for each scenario before evaluation.

All four controllers are evaluated under the same arena size, agent count,
rollout horizon, random non-overlapping initialization, overlap-safe dynamics,
and summary pipeline. For deterministic controllers, evaluation is performed
directly over five seeds. For MAPPO, each scenario is first trained in the same
PettingZoo-compatible environment and is then evaluated over five rollout
seeds using the same task metrics as SPIN.

\section{System Evaluation}

We evaluate the proposed framework in a custom PettingZoo-compatible discrete-time
multi-agent simulation environment in Python. The objective of this evaluation is not
intended to claim a production-ready decentralized control system or hardware-level
validation, but rather to test whether the core coordination mechanism of SPIN
produces coherent swarm-level behavior under constrained local computation in a
controlled coordination-focused simulation setting. In particular, the implementation
tests whether a lightweight perceptual prior, a bounded policy-weight reweighting
update, and clique-consistent tensor-network interaction constraints are sufficient
to support coherent behavior in three settings: motion toward a moving target, bounded
dispersion / area coverage, and decentralized coordination under multiple randomly
placed goals.

\subsection{System Settings}

The implementation used in this section executes the SPIN control loop described
above. All experiments are conducted in a bounded $100 \times 100$ continuous
arena with $N=10$ agents over $T=120$ synchronized control intervals under the
default experimental setting. Each agent maintains a local five-dimensional
behavioral state corresponding to the action basis.

$$\{\text{north}, \text{south}, \text{east}, \text{west}, \text{pinpoint}\}.$$

The sensing radius is fixed to $R_{\text{sense}} = 15.0$, the capped local
interaction graph retains at most $k_{\max}=8$ mutual nearest neighbors per
agent, the reweighting gain clamp is fixed to $\gamma = 5.0$, and the continuous
execution layer uses an action scale of $1.35$, a velocity damping factor of
$0.72$, and a maximum speed cap of $2.6$. Unless otherwise stated, the reported
hyperparameters were selected through lightweight empirical tuning to obtain stable
bounded execution across the three evaluation regimes, rather than through exhaustive
grid search. The same default settings were then held fixed across the reported SPIN
experiments.

The perceptual front-end is instantiated as a compact two-layer MLP with input
dimension $2$, hidden width $8$, and output dimension $5$. This network is
trained offline for $5000$ synthetic epochs using a cross-entropy objective over
hand-constructed directional targets derived from the relative target vector.
After pre-training, the network weights are frozen and each agent performs only
a forward pass at runtime, consistent with the low-overhead motivation of the
framework. For SPIN, APF-Velocity, and Distributed Auction-CBBA, the shared
PettingZoo-style simulator returns zero scalar rewards; the resulting
trajectories are produced entirely by the runtime control law rather than by
online reward optimization. The MAPPO baseline differs in this respect: during
training, it uses scenario-specific shaped rewards within the same environment
family, and is then evaluated under the same rollout metrics as the other
methods.

The implementation directly instantiates the analytical components introduced
earlier in the paper, including overlapping clique structure, clique-wise
tensor-network factorization, and marginal consistency across shared agents.
The topology module performs overlapping maximal-clique inference, the tensor
module builds clique MPS representations and computes exact one-site reduced
density matrices through left-right environment contractions, and the reweighting
update stage reconciles shared-agent reduced densities through an iterative
trace-distance consensus update before continuous-action execution. The main
remaining gap between theory and experiment therefore concerns empirical breadth
rather than the executable realization of the coordination mechanism itself. The
current experiments should be read as evidence that the SPIN simulator can realize
the intended coordination mechanism within a tractable PettingZoo-compatible
environment.

\subsection{Evaluation Protocol}

The simulation entrypoint evaluates three operating scenarios:
\begin{enumerate}
    \item \textbf{Tracking:} a single target moves on a randomized oscillatory
    path, and all agents are driven toward that target while maintaining a
    ring-like approach pattern;
    \item \textbf{Dispersion / Area Coverage:} the swarm starts from a random
    non-overlapping configuration and is guided by repulsion together with
    internal coverage anchors. The public evaluation metric is spatial entropy,
    with Voronoi area variance used as a complementary global coverage
    diagnostic;
    \item \textbf{Multi-Goal Coordination:} three random goals are placed in the
    arena, and each agent follows the nearest goal while occupying a local ring
    slot around that goal.
\end{enumerate}

At each control interval, the simulator executes four stages: (i) capped
adjacency construction from pairwise Euclidean distances, (ii) extraction of
overlapping maximal cliques from the capped local interaction graph, (iii)
per-agent bounded reweighting driven by the frozen perceptual network, and (iv)
clique-wise reduced-density computation with shared-agent reconciliation
followed by bounded motion execution through expected action vectors. The action
displacement magnitudes are fixed for the cardinal directions, while the
pinpoint behavior contributes a small micro-adjustment term.

\subsection{Validation Scope}

The current implementation supports three coordination-level claims under a
shared simulator interface; it is not intended as a strict equivalence test at
the low-level control-design layer. First, the perception-to-measure pipeline is
operational: a small frozen MLP can map a relative observation vector directly to
a normalized behavioral measure suitable for runtime execution. Second, the bounded
continuous measure update is operational: the bounded likelihood-ratio reweighting
step produces stable coordinate reshaping under repeated application, with explicit
state renormalization preventing numerical divergence. Third, the localized
coordination mechanism is operational at the clique-consistent level: overlapping
maximal cliques, clique-wise marginal matrix recovery, and shared-agent reconciliation
modify nearby agents' local action distributions in a way that changes swarm
organization across different tasks.

Taken together, these claims should be understood at the level of coordination
representation rather than low-level pursuit efficiency alone. The empirical goal
is not to show that SPIN is the strongest possible direct tracker, but that a
compact tensorized coordination layer can modulate executable motion behavior in
a stable, reusable, and task-flexible manner.

\section{Experimental Results}

\subsection{Observed Behavioral Regimes}
\label{subsec:behavior_results}

\begin{figure*}
    \centering
    \includegraphics[width=\textwidth]{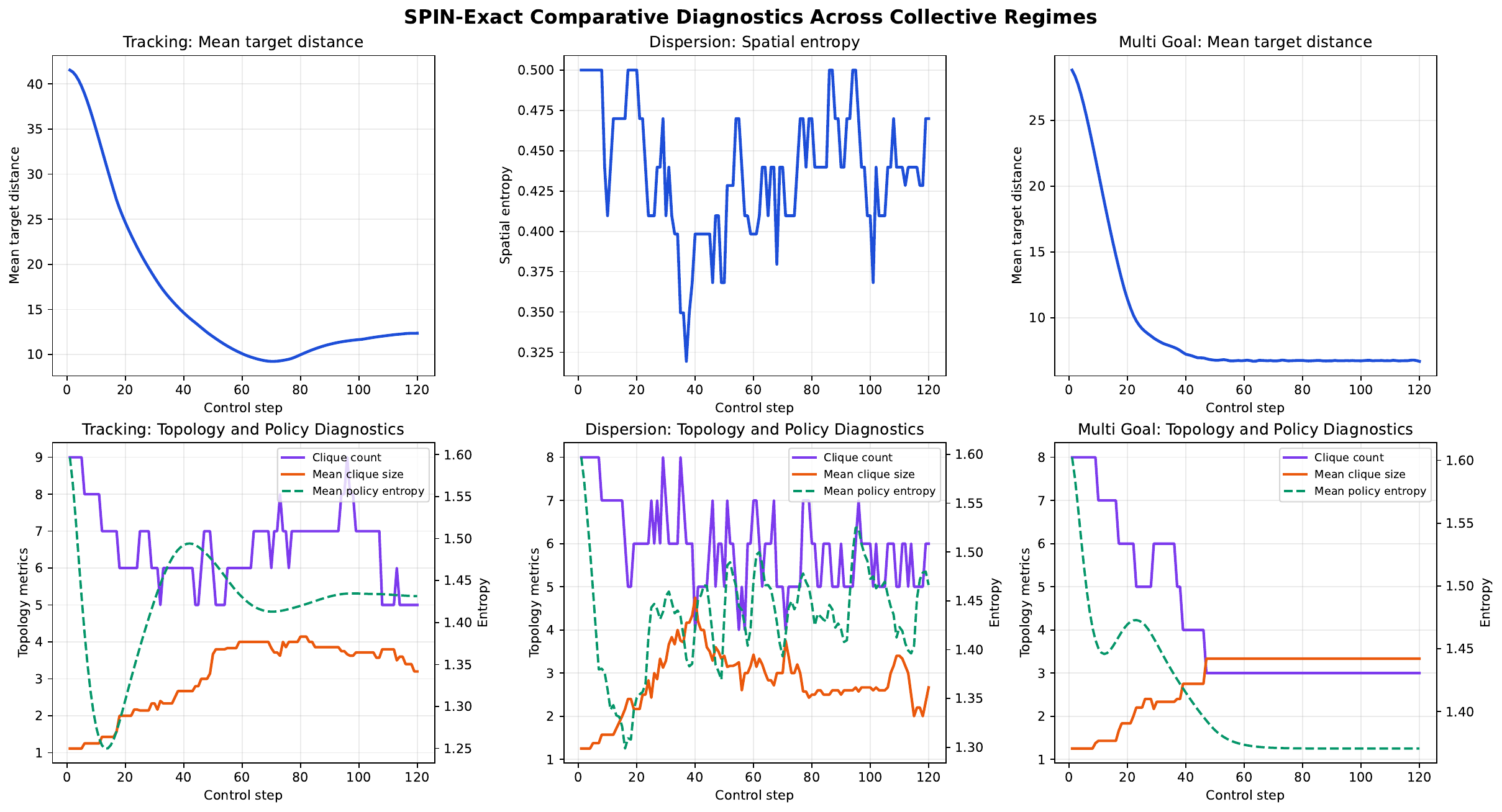}
    \caption{Comparative six-panel diagnostics for SPIN across Tracking,
    Dispersion / Area Coverage, and Multi-Goal Coordination. The top row
    reports the task-level evolution of each regime, showing mean target
    distance for Tracking and Multi-Goal Coordination and spatial entropy for
    Dispersion / Area Coverage. The bottom row reports topology and policy
    diagnostics, including clique count, mean clique size, and mean policy
    entropy. Together, these panels summarize the three collective regimes
    produced by the current control pipeline.}
    \label{fig:Fig.2}
\end{figure*}

To move beyond single-seed inspection, we executed five independent seeded trials
per scenario using the same simulator interface across SPIN, APF-Velocity,
Distributed Auction-CBBA, and MAPPO. The resulting summary statistics are reported
in Table~\ref{tab:trialsummary}, while Figure~\ref{fig:Fig.2} summarizes the
task-level, topology-level, and policy-level diagnostics of the three evaluated
coordination regimes. In particular, the figure reports the evolution of the public
task metric together with clique count, mean clique size, and mean policy entropy.

The regime-level diagnostics show that the same coordination pipeline produces
distinct collective organizations under different environmental objectives. The
Tracking regime converges toward the moving target while maintaining relatively low
policy entropy throughout the rollout. In the Dispersion / Area Coverage regime,
the swarm expands from a random non-overlapping initialization, reaching a final
spatial entropy of $0.452$, a mean trajectory length of $125.117$, and a final
Voronoi area variance of $705{,}625.535$, indicating bounded coverage regularization
rather than pure entropy maximization. The Multi-Goal Coordination regime partitions
the swarm among three randomly placed goals, producing stable decentralized
coordination across multiple target groups within the same shared coordination
mechanism.

\subsection{Comparison with Deterministic and Learned Baselines}

Table~\ref{tab:trialsummary} places SPIN alongside APF-Velocity, Distributed
Auction-CBBA, and MAPPO under matched rollout settings. This comparison clarifies the
intended role of SPIN in the benchmark. Because SPIN is evaluated here as a
coordination-layer mechanism rather than as a complete low-level controller, the
benchmark should be interpreted as a matched comparison of resulting coordinated
behavior under a shared simulator interface. Within that scope, the framework is not
optimized to dominate every specialized baseline in its strongest regime; rather, it
is designed to provide a reusable coordination layer that remains competitive across
qualitatively different swarm tasks without scenario-specific retraining. Accordingly,
the results are best understood as task-dependent trade-offs rather than evidence
of a single universally dominant method. APF-Velocity is the strongest pure tracking
specialist, reaching a final mean target distance of $4.621$ compared with SPIN's
$12.079$. In the Dispersion / Area Coverage regime, SPIN and APF-Velocity tie on the
public spatial-entropy metric ($0.452$), but SPIN reaches that final state with a
smaller mean trajectory length ($125.117$ versus $194.527$), indicating a more
conservative redistribution of the swarm. In the Multi-Goal Coordination regime,
SPIN and APF-Velocity achieve closely comparable final mean target distances
($6.746$ and $6.838$, respectively), while CBBA and MAPPO trail behind.

The learned MAPPO baseline remains useful because it tests whether an end-to-end
reward-driven policy can outperform the structured zero-shot controller when it
is allowed scenario-specific training. Under the converged MAPPO evaluation used
here, MAPPO becomes competitive in Tracking ($5.930$ final mean target distance)
but remains weaker in Dispersion / Area Coverage ($0.500$ final spatial entropy)
and clearly less effective in Multi-Goal Coordination ($15.066 \pm 9.628$ final
mean target distance). These results support the interpretation of SPIN as a
reusable coordination layer: it does not dominate every specialist baseline, but it
remains competitive across all three regimes without retraining.

\begin{table*}[t]
    \centering
    \caption{Five-trial summary statistics for SPIN and three baselines under
    matched random non-overlapping initialization, arena size, agent count, and
    rollout horizon. Tracking and Multi-Goal Coordination are scored by mean
    target distance, while Dispersion / Area Coverage is scored by spatial
    entropy. The $\Delta \pm \sigma$ column reports mean improvement and trial
    standard deviation. The final column reports mean trajectory length per agent
    over the full rollout.}
    \label{tab:trialsummary}
    \small
    \setlength{\tabcolsep}{4pt}
    \resizebox{0.9\textwidth}{!}{%
    \begin{tabular}{l l l c c c c c c}
        \hline
        \textbf{Scenario} & \textbf{Method} & \textbf{Init.}
        & \textbf{Final} & \textbf{$\Delta \pm \sigma$} & \textbf{Policy Ent.}
        & \textbf{Spatial Ent.} & \textbf{Voronoi Var.} & \textbf{Path Len.} \\
        \hline
        \multirow{4}{*}{\textbf{Tracking}} 
        & SPIN & 38.648 & 12.079 & 26.569 $\pm$ 5.400 & 1.453 & 0.372 & 663,049.351 & 63.394 \\
        & APF & 38.648 & 4.621 & 34.027 $\pm$ 4.302 & 0.694 & 0.266 & 791,072.428 & 84.424 \\
        & CBBA & 38.648 & 12.789 & 25.859 $\pm$ 4.532 & 0.368 & 0.452 & 142,142.951 & 130.723 \\
        & MAPPO & 38.648 & 5.930 & 32.718 $\pm$ 4.177 & 0.561 & 0.256 & 1,177,536.840 & 112.627 \\
        \hline
        \multirow{4}{*}{\textbf{Dispersion}} 
        & SPIN & 0.500 & 0.452 & -0.048 $\pm$ 0.036 & 1.435 & 0.452 & 705,625.535 & 125.117 \\
        & APF & 0.500 & 0.452 & -0.048 $\pm$ 0.041 & 0.823 & 0.452 & 612,027.231 & 194.527 \\
        & CBBA & 0.500 & 0.494 & -0.006 $\pm$ 0.012 & 0.245 & 0.494 & 244,542.355 & 100.588 \\
        & MAPPO & 0.500 & 0.500 & 0.000 $\pm$ 0.000 & 1.292 & 0.500 & 375,155.519 & 9.559 \\
        \hline
        \multirow{4}{*}{\textbf{Multi-Goal}} 
        & SPIN & 26.209 & 6.746 & 19.462 $\pm$ 3.005 & 1.370 & 0.476 & 275,328.877 & 37.588 \\
        & APF & 26.209 & 6.838 & 19.371 $\pm$ 2.966 & 0.685 & 0.482 & 275,679.708 & 111.514 \\
        & CBBA & 26.209 & 8.133 & 18.075 $\pm$ 3.066 & 0.174 & 0.488 & 235,052.587 & 108.919 \\
        & MAPPO & 26.209 & 15.066 & 11.143 $\pm$ 9.628 & 0.588 & 0.297 & 1,400,682.822 & 111.963 \\
        \hline
    \end{tabular}%
    }
\end{table*}

\subsection{Repeated-Trial Summary}

The repeated-trial statistics reinforce the regime-level trends shown in
Figures~\ref{fig:Fig.2}. Tracking reveals a clear specialist-versus-reusability
contrast: APF-Velocity is the strongest pure tracker, MAPPO is also competitive
after scenario-specific training, and SPIN remains functional but less aggressive.
Dispersion / Area Coverage shows that SPIN and APF-Velocity reach the same final
public entropy score while doing so with very different motion budgets. Multi-Goal
Coordination is the regime in which SPIN is strongest: it achieves a highly
competitive final mean target distance while also using the shortest average path
length among all four methods. The added mean-trajectory-length diagnostic is
particularly informative here because it distinguishes controllers that achieve a
favorable endpoint through compact structured motion from those that do so only after
much larger global travel.

\subsection{Tensor-Compression Validation and Scalability}
\label{subsec:tensor_scaling}

\begin{figure*}
    \centering
    \includegraphics[width=\textwidth]{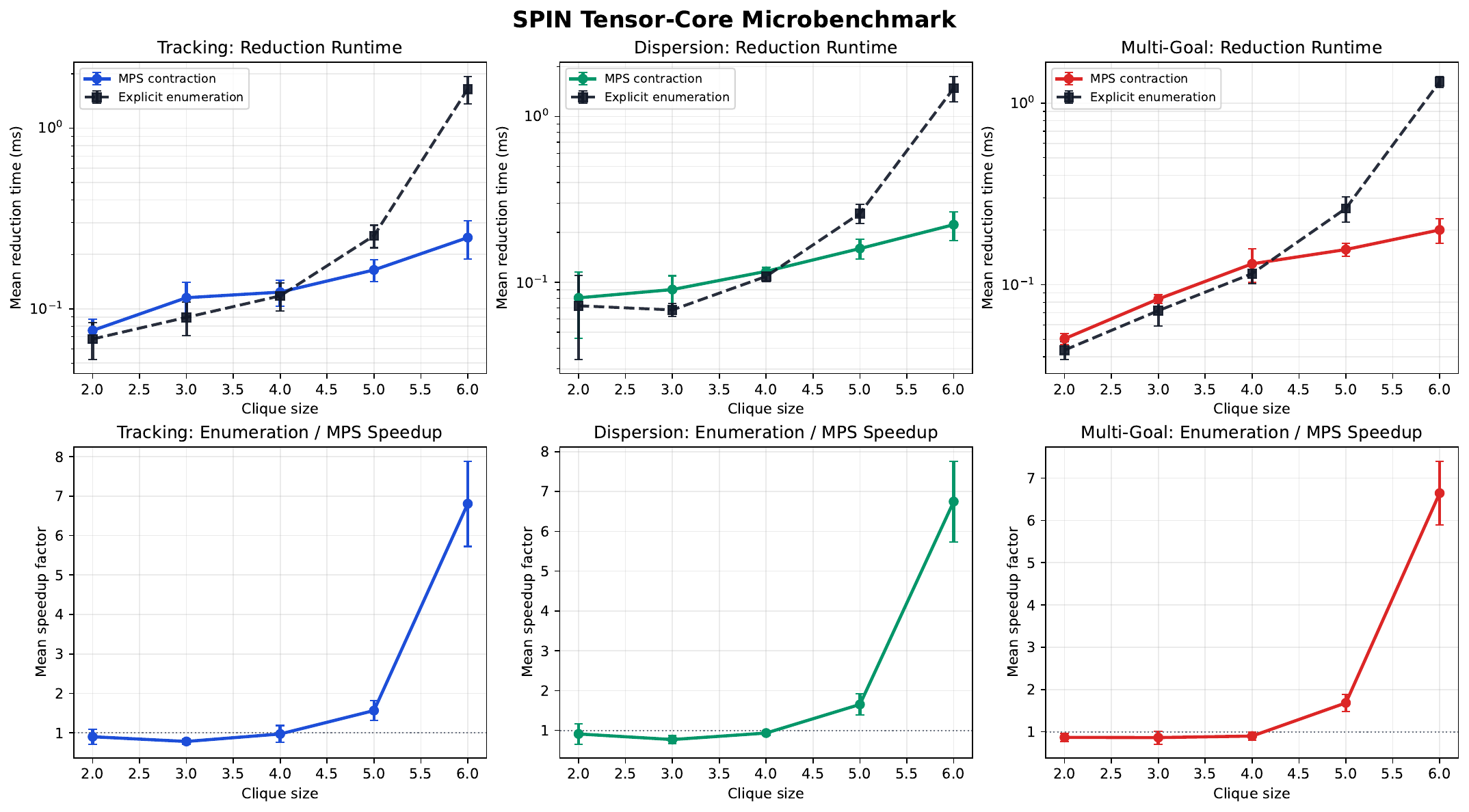}
    \caption{Tensor-core microbenchmark comparing the implemented MPS-based
    clique reduction against explicit joint-state enumeration for clique sizes
    2--6 across the three evaluated regimes. Top row: mean reduction runtime.
    Bottom row: enumeration-to-MPS speedup. The two methods are comparable at
    very small clique sizes, but the compressed contraction becomes
    increasingly favorable as clique size grows.}
    \label{fig:Fig.3}
\end{figure*}

\begin{figure*}
    \centering
    \includegraphics[width=0.9\textwidth]{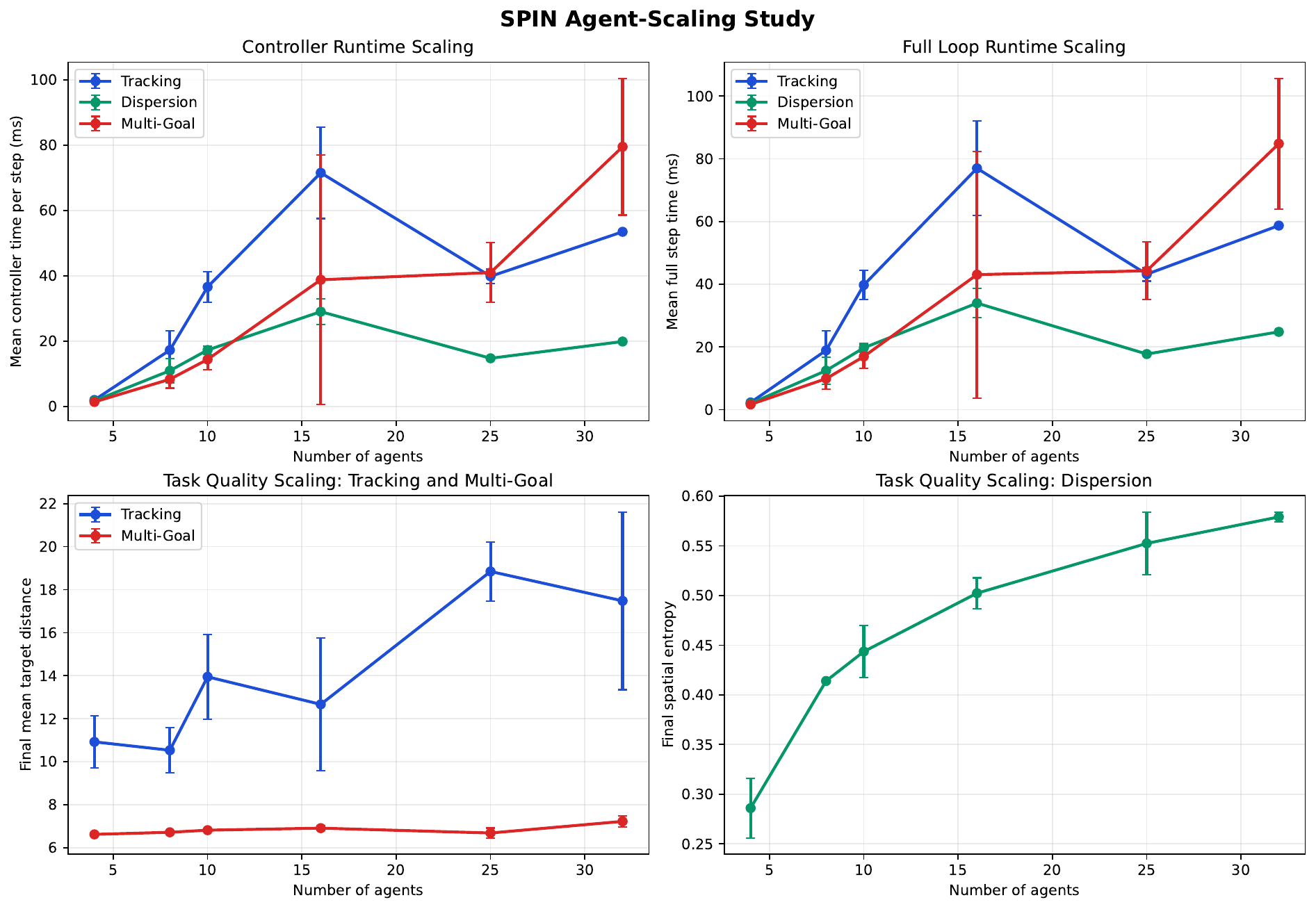}
    \caption{System-level agent scaling study for
    $N \in \{4,8,10,16,25,32\}$. The top row reports mean controller-step
    runtime and full-loop runtime. The bottom row reports final task metrics:
    mean target distance for Tracking and Multi-Goal Coordination, and spatial
    entropy for Dispersion / Area Coverage.}
    \label{fig:Fig.4}
\end{figure*}

This subsection serves as a targeted component-level validation of SPIN's
tensor-compression mechanism by comparing clique-wise MPS contraction against
explicit joint-state enumeration and by measuring the resulting scaling behavior
as local interaction size increases. Computational scalability was evaluated from
both algorithmic and system perspectives. The tensor-core microbenchmark compares
the proposed MPS contraction with explicit joint-state enumeration for clique sizes
$\kappa=2$--$6$ across the three coordination regimes. As shown in
Figure~\ref{fig:Fig.3}, both methods exhibit comparable runtime for small cliques,
whereas the MPS implementation becomes progressively faster as clique size
increases. This directly supports the computational motivation for tensorized clique
compression as a local coordination mechanism under growing interaction complexity.

End-to-end scalability was evaluated for $N\in\{4,8,10,16,25,32\}$ over a fixed
120-step horizon. Figure~\ref{fig:Fig.4} reports controller-step runtime, total
runtime, and final task metrics. Runtime increases with swarm size while task
performance remains stable, indicating that the computational cost of the full
loop grows without immediate collapse of coordination quality. A larger stress
test with $N\in\{25,50,75,100\}$ is summarized in Table~\ref{tab:large_scale_stress},
reporting tracking distance, dispersion entropy, multi-goal distance, and
distribution statistics (P90 and MAD) for SPIN, APF-Velocity, and Distributed
Auction-CBBA.

\begin{table*}[t]
\centering
\caption{Large-scale crowded-swarm stress test for SPIN, APF-Velocity, and
Distributed Auction-CBBA over $N \in \{25,50,75,100\}$ agents and 120 control
steps. Values report mean $\pm$ standard deviation over three seeds. Tracking
and Multi-Goal Coordination use final mean target distance, where lower is
better. Dispersion / Area Coverage uses final spatial entropy, where higher is
better. P90 and MAD report the 90th-percentile and mean absolute deviation of
the final per-agent target-distance distribution. Best primary task value per
agent count and regime is bolded.}
\label{tab:large_scale_stress}
\scriptsize
\setlength{\tabcolsep}{3.5pt}
\resizebox{\textwidth}{!}{
\begin{tabular}{llccccccc}
\hline
$N$ & Method
& Track dist. $\downarrow$
& Track P90 $\downarrow$
& Track MAD $\downarrow$
& Disp. ent. $\uparrow$
& Multi dist. $\downarrow$
& Multi P90 $\downarrow$
& Multi MAD $\downarrow$ \\
\hline
25 & SPIN & 18.846 $\pm$ 1.380 & 27.76 & 5.78 & 0.552 $\pm$ 0.031 & \textbf{6.684 $\pm$ 0.232} & 7.94 & 1.17 \\
25 & APF & \textbf{10.214 $\pm$ 0.047} & 12.13 & 1.65 & 0.504 $\pm$ 0.053 & 7.255 $\pm$ 0.080 & 7.76 & 0.34 \\
25 & CBBA & 16.268 $\pm$ 0.123 & 19.08 & 2.70 & \textbf{0.622 $\pm$ 0.017} & 10.759 $\pm$ 0.048 & 13.29 & 1.89 \\
\hline
50 & SPIN & 19.010 $\pm$ 3.134 & 28.17 & 5.95 & 0.633 $\pm$ 0.005 & \textbf{7.647 $\pm$ 0.108} & 10.45 & 1.64 \\
50 & APF & \textbf{11.252 $\pm$ 0.047} & 14.93 & 2.72 & 0.566 $\pm$ 0.029 & 8.029 $\pm$ 0.078 & 10.12 & 1.40 \\
50 & CBBA & 20.429 $\pm$ 0.039 & 25.40 & 4.04 & \textbf{0.694 $\pm$ 0.015} & 13.787 $\pm$ 0.227 & 17.93 & 2.96 \\
\hline
75 & SPIN & 17.191 $\pm$ 0.584 & 27.30 & 6.65 & 0.700 $\pm$ 0.010 & \textbf{8.132 $\pm$ 0.249} & 11.61 & 2.27 \\
75 & APF & \textbf{12.300 $\pm$ 0.185} & 18.02 & 3.74 & 0.645 $\pm$ 0.011 & 8.522 $\pm$ 0.049 & 10.97 & 1.85 \\
75 & CBBA & 23.657 $\pm$ 0.047 & 30.44 & 5.12 & \textbf{0.774 $\pm$ 0.007} & 15.999 $\pm$ 0.172 & 21.20 & 3.55 \\
\hline
100 & SPIN & 19.071 $\pm$ 1.858 & 31.15 & 7.53 & 0.740 $\pm$ 0.010 & 8.986 $\pm$ 0.099 & 13.46 & 2.92 \\
100 & APF & \textbf{13.428 $\pm$ 0.318} & 19.85 & 4.24 & 0.693 $\pm$ 0.004 & \textbf{8.976 $\pm$ 0.073} & 12.12 & 2.32 \\
100 & CBBA & 26.310 $\pm$ 0.150 & 34.37 & 5.93 & \textbf{0.811 $\pm$ 0.009} & 18.058 $\pm$ 0.008 & 23.84 & 4.17 \\
\hline
\end{tabular}
}
\end{table*}

\subsection{Perception Robustness}
\label{subsec:robustness}

\begin{figure*}
    \centering
    \includegraphics[width=0.85\textwidth]{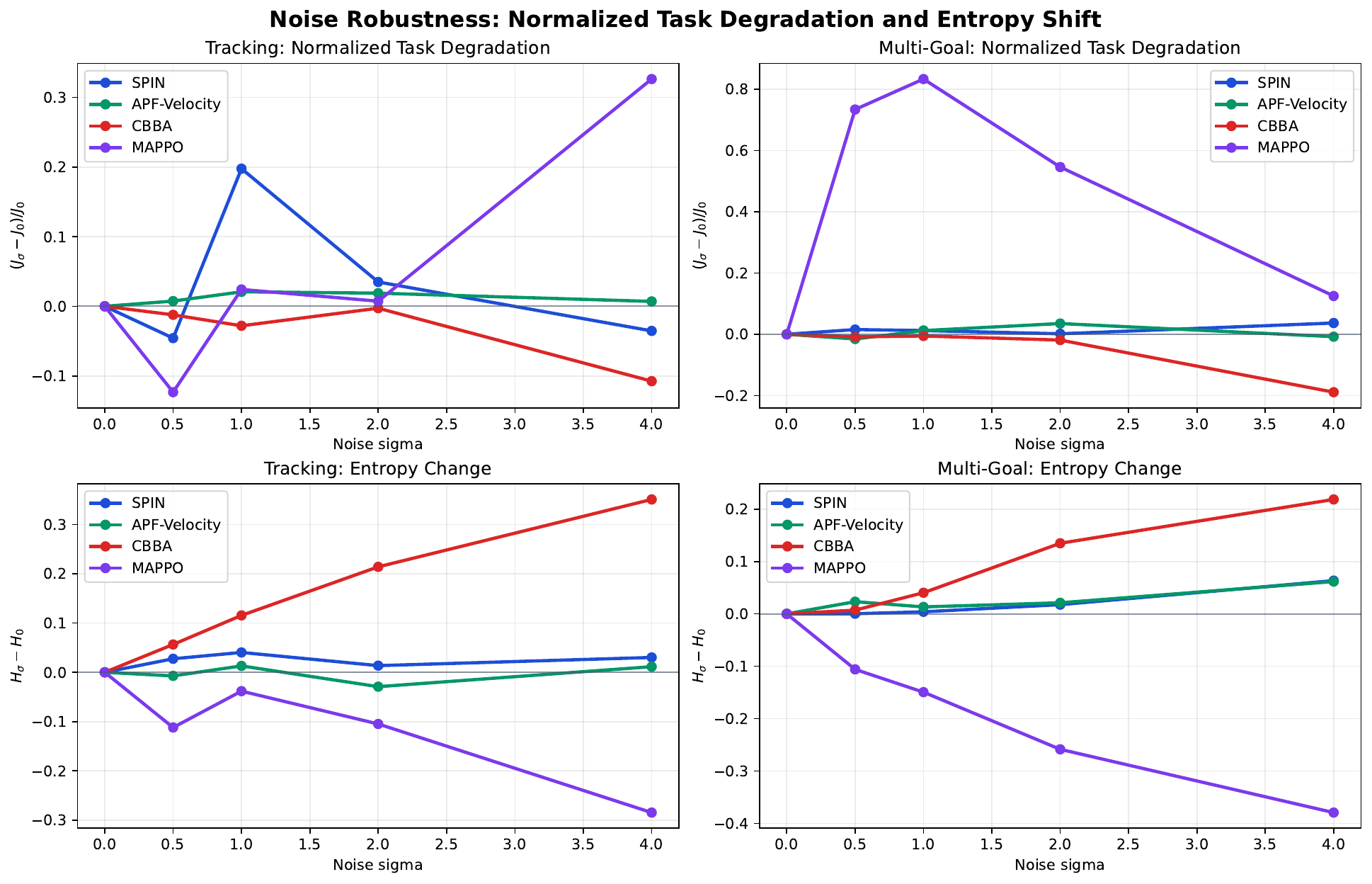}
    \caption{Normalized degradation under Gaussian observation/descriptor noise.
    Curves show mean performance over five seeds at each noise level. The
    response is plotted only at measured noise levels without interpolation,
    since several controllers exhibit non-monotonic sensitivity to perturbations.
    The top row reports normalized task degradation relative to each method's
    own noise-free baseline, while the bottom row reports the corresponding
    entropy shift.}
    \label{fig:Fig.5}
\end{figure*}

Perception robustness was evaluated using two complementary stress tests on the
Tracking and Multi-Goal Coordination tasks. These experiments probe the sensitivity
of each controller to degraded local information rather than full sensor or
hardware validation. In the first, zero-mean Gaussian noise was injected into each
controller's observations (or SPIN's policy signal) with standard deviation
$\sigma\in\{0.0,0.5,1.0,2.0,4.0\}$, while leaving the simulator state unchanged.
Figure~\ref{fig:Fig.5} reports normalized performance degradation relative to each
method's noise-free baseline together with the corresponding entropy change.

\begin{figure*}
    \centering
    \includegraphics[width=0.85\textwidth]{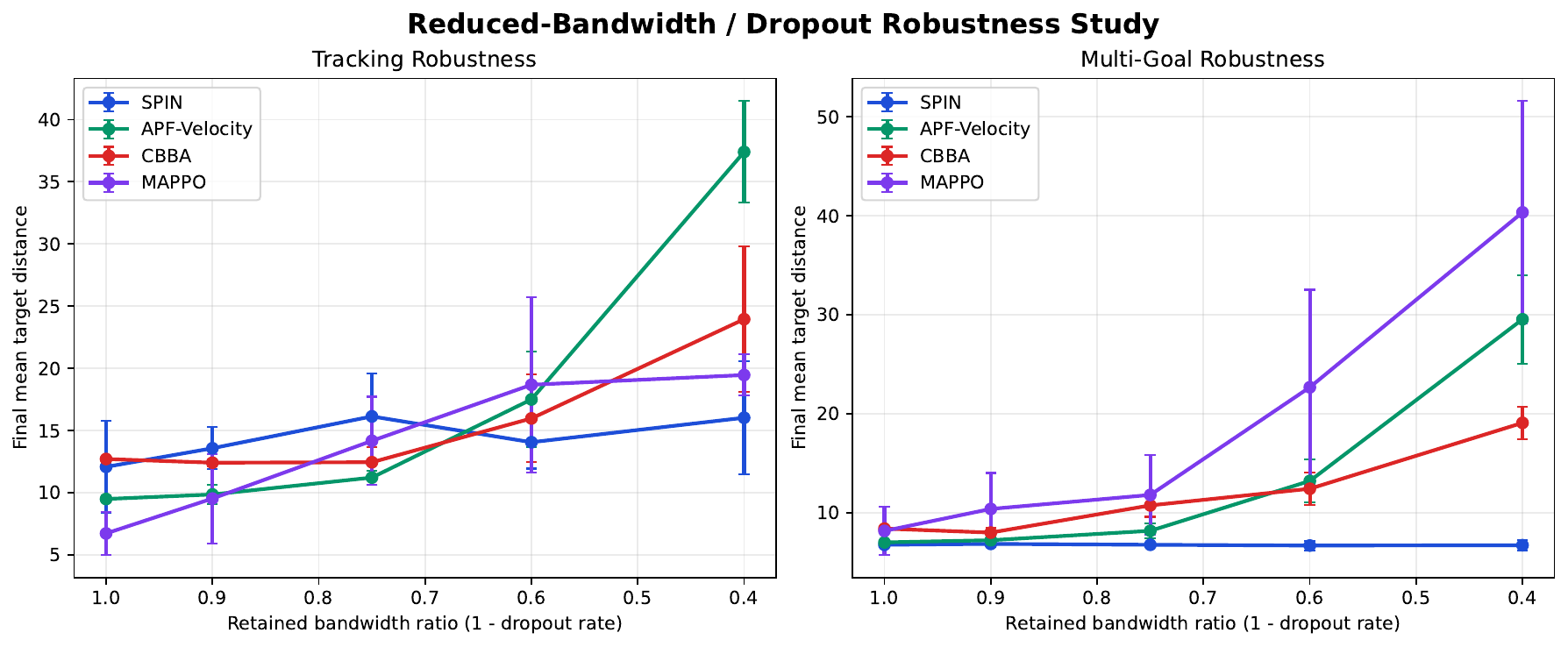}
    \caption{Reduced-bandwidth / dropout robustness study on Tracking and
    Multi-Goal Coordination. Each point reports the mean final mean target
    distance over five seeds at retained bandwidth ratios
    $1-r \in \{1.0, 0.9, 0.75, 0.6, 0.4\}$ corresponding to dropout rates
    $r \in \{0.0, 0.1, 0.25, 0.4, 0.6\}$. Error bars denote one standard
    deviation across trials.}
    \label{fig:Fig.6}
\end{figure*}

The second experiment evaluates robustness to partial information loss using
coordinate-wise dropout with rates $r\in\{0.0,0.1,0.25,0.4,0.6\}$, corresponding
to retained bandwidth ratios $1-r$. Dropout is applied to the same controller
inputs while the environment remains unchanged. Figure~\ref{fig:Fig.6} reports
the final mean target distance over five trials for each tested dropout level.

\section{Discussion}
\label{discussion}

The results in Section~\ref{subsec:behavior_results} suggest that SPIN should be
understood primarily as a coordination-level mechanism rather than as a specialized
controller for any single task. Across Tracking, Dispersion / Area Coverage, and
Multi-Goal Coordination, the same underlying coordination pipeline produces distinct
collective behaviors without requiring scenario-specific retraining or redesign of
the coordination logic itself. In this sense, the main contribution of SPIN is not
that it universally dominates all baselines, but that it provides a reusable
structured coordination layer through which different swarm-level regimes can emerge
from the same localized tensorized interaction mechanism.

This interpretation is most clearly supported by the contrast among the three basic
evaluated regimes. In Tracking, the swarm evolves toward an ordered high-consensus
configuration around a shared moving objective. In Dispersion / Area Coverage, the
same coordination machinery instead yields bounded coverage regularization, rather
than pure entropy maximization, under distributed repulsion and local anchors. In
Multi-Goal Coordination, the swarm separates into multiple locally coordinated groups
while still operating within the same shared inference pipeline. Taken together,
these patterns indicate that SPIN is capable of modulating collective organization
through local coordination structure, rather than through task-specific controller
redesign.

The comparative baseline results further clarify the role of the framework.
APF-Velocity remains the strongest specialized tracker, while Distributed Auction-CBBA
is more naturally aligned with dispersion-oriented objectives. SPIN is most compelling
in Multi-Goal Coordination, where it achieves a final mean target distance comparable
to the strongest competing baseline together with the shortest average path length
among the compared methods. This combination is important because it suggests not only
successful endpoint coordination, but also compact and structured swarm motion in
reaching that outcome. Accordingly, the empirical picture emerging from the current
benchmark is not one of universal dominance, but of task-dependent trade-offs in
which SPIN is strongest when coordination demands become more locally coupled,
distributed, and structurally heterogeneous.

The computational results in Section~\ref{subsec:tensor_scaling} reinforce this
interpretation. At the tensor-core level, clique-wise MPS reduction becomes
increasingly favorable as clique size grows, supporting the original motivation
for tensorized compression of localized joint coordination structure. At the
system level, runtime growth is dominated by the full coordination loop rather
than tensor contraction alone, yet the framework remains tractable under
increasing swarm size and under the larger crowded-swarm stress tests. These
results suggest that the practical value of SPIN lies not only in abstract
representation, but in preserving executable localized coordination under rising
interaction density without reverting to explicit joint-state enumeration or
scenario-specific online optimization. Accordingly, the present study does provide
component-level evidence for the tensorized compression mechanism, even though it
does not exhaustively remove every module in a full end-to-end ablation suite.

The robustness experiments in Section~\ref{subsec:robustness} provide a further view
of the framework from the perspective of degraded local information. Gaussian
perturbations probe sensitivity to noisy observations, while coordinate dropout
probes resilience to partial information loss. Since all controllers are evaluated
under identical test-time perturbation schedules, the resulting trends should be read
as relative robustness comparisons rather than as absolute sensor-validation claims.
Within that scope, the experiments show that SPIN maintains coherent behavior under
controlled degradation while preserving its strongest relative advantage in the more
coordination-intensive regimes. Overall, the present results support the
interpretation of SPIN as a decentralized coordination framework with favorable
scaling behavior over the evaluated regimes whose principal utility lies in
structured multi-agent organization under complex local interaction, rather than in
optimization for a single narrow swarm objective.

\subsection{Limitations of the Current Evaluation}

The present study has several limitations. First, SPIN is a coordination-level
framework rather than a complete low-level control stack. Its executable behavior
therefore still depends on the geometric motion laws and environment-side actuation
mechanisms used to realize final movement, and future work should evaluate how SPIN
performs when coupled with different low-level control frameworks. Second, although
the current implementation could in principle be further optimized for particular
tasks, the experiments in this paper intentionally use a conceptually aligned and
comparatively uniform implementation in order to preserve fairness across baseline
comparisons, rather than a task-specialized or production-oriented realization.

Moreover, due to resource constraints, the current evaluation does not include a
real-world implementation of SPIN-based coordination. Finally, in order to make the
paper tight and neat, the experimental section presents a selected set of
representative studies rather than a fully comprehensive empirical investigation.
Accordingly, the present evaluation should be interpreted as evidence of coordinated
navigation quality and interaction-level behavior, rather than as a full validation
of collision safety or deployment readiness. These limitations should therefore be
understood as constraints on empirical breadth and deployment scope, rather than as
contradictions of the coordination-level claims developed in this paper.

\section{Conclusion}

This paper introduced SPIN as a tensor-network-based framework for decentralized
local coordination that mitigates interaction complexity through clique-level
tensorized compression. The central idea is to decouple perception from
coordination: a lightweight perceptual network is trained offline to map local
observations into target measures, while runtime behavior is governed by bounded
likelihood-ratio reweighting, localized state-vector updates, and
tensor-mediated structural interactions. This design yields a coordination
pipeline in which behavioral adaptation is performed through direct algebraic
state updates rather than online optimization.

Within the present simulation setting, the same SPIN engine was shown to support
three qualitatively distinct collective regimes: ordered single-target tracking,
dispersion / area coverage, and structured multi-goal coordination. The results
indicate that SPIN is most compelling in multi-goal, high-density, and locally
coupled coordination settings. They further suggest that the tensorized structure of
SPIN provides favorable computational scaling behavior for clique-level coordination
over the tested clique sizes and swarm sizes, even though such behavior should not
be interpreted as a universal scalability guarantee beyond the evaluated regimes.
Accordingly, the framework is better understood as a reusable coordination layer
for swarm organization than as a single-purpose tracking controller.

Future work should focus on tightening the bridge between theory and
implementation through stronger coverage-oriented control, broader repeated-trial
evaluation, and expanded baseline and training-budget comparisons. Nevertheless,
the present results already demonstrate that the core SPIN design principles
admit a bounded and interpretable executable prototype.

\bibliographystyle{plain}
\bibliography{references}

@article{sheng2006distributed,
  title={Distributed multi-robot coordination in area exploration},
  author={Sheng, Weihua and Yang, Qingyan and Tan, Jindong and Xi, Ning},
  journal={Robotics and autonomous systems},
  volume={54},
  number={12},
  pages={945--955},
  year={2006},
  publisher={Elsevier}
}

@inproceedings{chung2006decentralized,
  title={A decentralized motion coordination strategy for dynamic target tracking},
  author={Chung, Timothy H and Burdick, Joel W and Murray, Richard M},
  booktitle={Proceedings 2006 IEEE International Conference on Robotics and Automation, 2006. ICRA 2006.},
  pages={2416--2422},
  year={2006},
  organization={IEEE}
}

@article{brambilla2013swarm,
  title={Swarm robotics: a review from the swarm engineering perspective},
  author={Brambilla, Manuele and Ferrante, Eliseo and Birattari, Mauro and Dorigo, Marco},
  journal={Swarm Intelligence},
  volume={7},
  number={1},
  pages={1--41},
  year={2013},
  publisher={Springer}
}

@article{vinyals2019grandmaster,
  title={Grandmaster level in StarCraft II using multi-agent reinforcement learning},
  author={Vinyals, Oriol and Babuschkin, Igor and Czarnecki, Wojciech M and Mathieu, Micha{\"e}l and Dudzik, Andrew and Chung, Junyoung and Choi, David H and Powell, Richard and Ewalds, Timo and Georgiev, Petko and others},
  journal={nature},
  volume={575},
  number={7782},
  pages={350--354},
  year={2019},
  publisher={Nature Publishing Group UK London}
}

@article{bekmezci2013flying,
  title={Flying ad-hoc networks (FANETs): A survey},
  author={Bekmezci, Ilker and Sahingoz, Ozgur Koray and Temel, {\c{S}}amil},
  journal={Ad Hoc Networks},
  volume={11},
  number={3},
  pages={1254--1270},
  year={2013},
  publisher={Elsevier}
}

@article{hayat2016survey,
  title={Survey on unmanned aerial vehicle networks for civil applications: A communications viewpoint},
  author={Hayat, Samira and Yanmaz, Ev{\c{s}}en and Muzaffar, Raheeb},
  journal={IEEE communications surveys \& tutorials},
  volume={18},
  number={4},
  pages={2624--2661},
  year={2016},
  publisher={IEEE}
}

@article{rashid2020monotonic,
  title={Monotonic value function factorisation for deep multi-agent reinforcement learning},
  author={Rashid, Tabish and Samvelyan, Mikayel and De Witt, Christian Schroeder and Farquhar, Gregory and Foerster, Jakob and Whiteson, Shimon},
  journal={Journal of Machine Learning Research},
  volume={21},
  number={178},
  pages={1--51},
  year={2020}
}

@book{pearl2014probabilistic,
  title={Probabilistic reasoning in intelligent systems: networks of plausible inference},
  author={Pearl, Judea},
  year={2014},
  publisher={Elsevier}
}

@inproceedings{ihler2004nonparametric,
  title={Nonparametric belief propagation for self-calibration in sensor networks},
  author={Ihler, Alexander T and Fisher III, John W and Moses, Randolph L and Willsky, Alan S},
  booktitle={Proceedings of the 3rd international symposium on Information processing in sensor networks},
  pages={225--233},
  year={2004}
}

@article{mozaffari2019tutorial,
  title={A tutorial on UAVs for wireless networks: Applications, challenges, and open problems},
  author={Mozaffari, Mohammad and Saad, Walid and Bennis, Mehdi and Nam, Young-Han and Debbah, M{\'e}rouane},
  journal={IEEE communications surveys \& tutorials},
  volume={21},
  number={3},
  pages={2334--2360},
  year={2019},
  publisher={IEEE}
}

@inproceedings{svoboda2022deep,
  title={Deep learning on microcontrollers: A study on deployment costs and challenges},
  author={Svoboda, Filip and Fernandez-Marques, Javier and Liberis, Edgar and Lane, Nicholas D},
  booktitle={Proceedings of the 2nd European Workshop on Machine Learning and Systems},
  pages={54--63},
  year={2022}
}

@article{li2017learning,
  title={Learning without forgetting},
  author={Li, Zhizhong and Hoiem, Derek},
  journal={IEEE transactions on pattern analysis and machine intelligence},
  volume={40},
  number={12},
  pages={2935--2947},
  year={2017},
  publisher={IEEE}
}

@article{battaglia2018relational,
  title={Relational inductive biases, deep learning, and graph networks},
  author={Battaglia, Peter W and Hamrick, Jessica B and Bapst, Victor and Sanchez-Gonzalez, Alvaro and Zambaldi, Vinicius and Malinowski, Mateusz and Tacchetti, Andrea and Raposo, David and Santoro, Adam and Faulkner, Ryan and others},
  journal={arXiv preprint arXiv:1806.01261},
  year={2018}
}

@book{busemeyer2012quantum,
  title={Quantum models of cognition and decision},
  author={Busemeyer, Jerome R and Bruza, Peter D},
  year={2012},
  publisher={Cambridge University Press}
}

@inproceedings{sun2004particle,
  title={Particle swarm optimization with particles having quantum behavior},
  author={Sun, Jun and Feng, Bin and Xu, Wenbo},
  booktitle={Proceedings of the 2004 congress on evolutionary computation (IEEE Cat. No. 04TH8753)},
  volume={1},
  pages={325--331},
  year={2004},
  organization={IEEE}
}

@article{verbraeken2020survey,
  title={A survey on distributed machine learning},
  author={Verbraeken, Joost and Wolting, Matthijs and Katzy, Jonathan and Kloppenburg, Jeroen and Verbelen, Tim and Rellermeyer, Jan S},
  journal={Acm computing surveys (csur)},
  volume={53},
  number={2},
  pages={1--33},
  year={2020},
  publisher={Acm New York, NY, USA}
}

@article{foerster2016learning,
  title={Learning to communicate with deep multi-agent reinforcement learning},
  author={Foerster, Jakob and Assael, Ioannis Alexandros and De Freitas, Nando and Whiteson, Shimon},
  journal={Advances in neural information processing systems},
  volume={29},
  year={2016}
}

@inproceedings{das2019tarmac,
  title={Tarmac: Targeted multi-agent communication},
  author={Das, Abhishek and Gervet, Th{\'e}ophile and Romoff, Joshua and Batra, Dhruv and Parikh, Devi and Rabbat, Mike and Pineau, Joelle},
  booktitle={International Conference on machine learning},
  pages={1538--1546},
  year={2019},
  organization={PMLR}
}

@article{yu2022surprising,
  title={The surprising effectiveness of ppo in cooperative multi-agent games},
  author={Yu, Chao and Velu, Akash and Vinitsky, Eugene and Gao, Jiaxuan and Wang, Yu and Bayen, Alexandre and Wu, Yi},
  journal={Advances in neural information processing systems},
  volume={35},
  pages={24611--24624},
  year={2022}
}

@inproceedings{schulman2015trust,
  title={Trust region policy optimization},
  author={Schulman, John and Levine, Sergey and Abbeel, Pieter and Jordan, Michael and Moritz, Philipp},
  booktitle={International conference on machine learning},
  pages={1889--1897},
  year={2015},
  organization={PMLR}
}

@article{schulman2017proximal,
  title={Proximal policy optimization algorithms},
  author={Schulman, John and Wolski, Filip and Dhariwal, Prafulla and Radford, Alec and Klimov, Oleg},
  journal={arXiv preprint arXiv:1707.06347},
  year={2017}
}

@article{li2020deep,
  title={Deep learning for lidar point clouds in autonomous driving: A review},
  author={Li, Ying and Ma, Lingfei and Zhong, Zilong and Liu, Fei and Chapman, Michael A and Cao, Dongpu and Li, Jonathan},
  journal={IEEE Transactions on Neural Networks and Learning Systems},
  volume={32},
  number={8},
  pages={3412--3432},
  year={2020},
  publisher={IEEE}
}

@book{koller2009probabilistic,
  title={Probabilistic graphical models: principles and techniques},
  author={Koller, Daphne and Friedman, Nir},
  year={2009},
  publisher={MIT press}
}

@article{wainwright2008graphical,
  title={Graphical models, exponential families, and variational inference},
  author={Wainwright, Martin J and Jordan, Michael I},
  journal={Foundations and Trends{\textregistered} in Machine Learning},
  volume={1},
  number={1-2},
  pages={1--305},
  year={2008},
  publisher={Emerald Publishing Limited}
}

@article{orus2014practical,
  title={A practical introduction to tensor networks: Matrix product states and projected entangled pair states},
  author={Or{\'u}s, Rom{\'a}n},
  journal={Annals of physics},
  volume={349},
  pages={117--158},
  year={2014},
  publisher={Elsevier}
}

@inproceedings{yu2022the,
  title={The Surprising Effectiveness of {PPO} in Cooperative Multi-Agent Games},
  author={Chao Yu and Akash Velu and Eugene Vinitsky and Jiaxuan Gao and Yu Wang and Alexandre Bayen and Yi Wu},
  booktitle={Thirty-sixth Conference on Neural Information Processing Systems Datasets and Benchmarks Track},
  year={2022}
}

\end{document}